# IoT-Based Air Quality Monitoring System with Machine Learning for Accurate and Real-time Data Analysis

Hemanth Karnati

*Abstract*: *Air pollution in urban areas has severe consequences for both human health and the environment, predominantly caused by exhaust emissions from vehicles. To address the issue of air pollution awareness, Air Pollution Monitoring systems are used to measure the concentration of gases like CO2, smoke, alcohol, benzene, and NH3 present in the air. However, current mobile applications are unable to provide users with real-time data specific to their location. In this paper, we propose the development of a portable air quality detection device that can be used anywhere. The data collected will be stored and visualized using the cloud-based web app ThinkSpeak. The device utilizes two sensors, MQ135 and MQ3, to detect harmful gases and measure air quality in parts per million (PPM). Additionally, machine learning analysis will be employed on the collected data.*

## INTRODUCTION

Air quality plays a crucial role in human health and the well-being of the environment. Unfortunately, air pollution has been on the rise due to various sources such as vehicle emissions, industrial activities, energy production, and natural disasters like wildfires. Understanding and assessing the quality of the air we breathe is of utmost importance. Air Quality Monitoring (AQM) systems, integrated with sensors and advanced technologies, are utilized to measure particulate matter and air pollutants like ozone, nitrogen oxides, and sulfur dioxide. The data collected by these systems helps formulate policies, monitor pollution reduction efforts, and empower the public to make informed decisions regarding their health and well-being.

Currently, AQM stations are primarily used for calculating the Air Quality Index (AQI) and monitoring pollution. However, the infrastructure requirements, operational complexities, and ongoing expenses associated with these stations limit the expansion of AQM networks and the availability of air pollution data. To overcome these limitations, it is imperative to develop low-cost, efficient, and real-time data-sensing devices. IoT technology provides a promising solution, with recent advancements allowing the use of IoT sensors in various domains, including smart cities, smart mobiles, smart refrigerators, and smartwatches. Leveraging IoT, air quality can be monitored remotely using sensors (e.g., temperature and pressure sensors, noise sensors), Arduino for data processing, and cloud platforms for storage. Machine learning algorithms, such as Linear Regression, Random Forest, XGBoost, and ARIMA models, have also proven effective in forecasting and predicting air pollutant levels. The availability of affordable sensors and data processing tools has enabled the deployment of air quality monitoring systems on a large scale. However, maintaining the accuracy of these systems is crucial, as erroneous data can lead to flawed policy decisions and ineffective mitigation efforts. Regular calibration and validation are essential to ensure the accuracy of air quality monitoring systems.

This paper presents several contributions:

- Development of a low-cost and user-friendly air pollution monitoring system.
- Real-time data gathering capabilities within the AQM system.
- Utilization of Blynk for real-time data visualization.
- Adoption of ThingSpeak, an open-source software, for day-to-day pollution visualization.

The importance of air quality and the necessity of monitoring systems are discussed, highlighting the limitations of current AQM stations and the need for cost-effective and efficient solutions using IoT technology. The paper outlines the use of IoT sensors, Arduino, cloud platforms, and machine learning algorithms for real-time air quality monitoring. The proposed system includes a low-cost, user-friendly AQM system capable of gathering real-time data, a website displaying the Air Quality Index, and the integration of Blynk to access IoT sensor data and Thingspeak for data visualization.

## PROPOSED WORK

The proposed framework for the IoT device with MQ3 and MQ135 sensors, NodeMCU processor, Arduino IDE, ThingSpeak, and Blynk platforms, consists of a well-structured hardware and software architecture, as well as an ML analysis and visualization component. This framework aims to provide a reliable and efficient system for collecting, analyzing, and visualizing air quality and alcohol level data in real-time.

Hardware Architecture: The hardware architecture of the IoT device includes the sensors, NodeMCU processor, WiFi module, power source, and other necessary components. The sensors, MQ3 and MQ135, are responsible for collecting air quality and alcohol level data respectively. The NodeMCU processor processes the collected data and controls the behavior of the device.

The WiFi module enables internet connectivity for the device to transfer the collected data to the ThingSpeak platform. The power source supplies power to the device.

Software Architecture: The software architecture of the IoT device is composed of several layers, including a sensor data acquisition layer, data processing layer, internet connectivity layer, cloud data storage layer, and user interface layer. The firmware running on the NodeMCU controls the device behavior and sends collected data to the ThingSpeak platform for storage and analysis.

The Blynk app provides a real-time display of the collected data to the user. ML Analysis and Visualization Component: After collecting data from the ThingSpeak platform, the next step is to analyze and visualize the collected data using ML algorithms.

Overall, this proposed framework for analyzing and visualizing the data collected using the IoT device with

MQ3 and MQ135 sensors, NodeMCU processor, Arduino IDE, ThingSpeak, and Blynk platforms provides a structured approach for processing, selecting features, selecting appropriate ML algorithms, building, and training models, evaluating models, and visualizing data. This framework can help to provide meaningful insights for decision-making in various applications, such as environmental monitoring and public.

# FLOW DIAGRAM

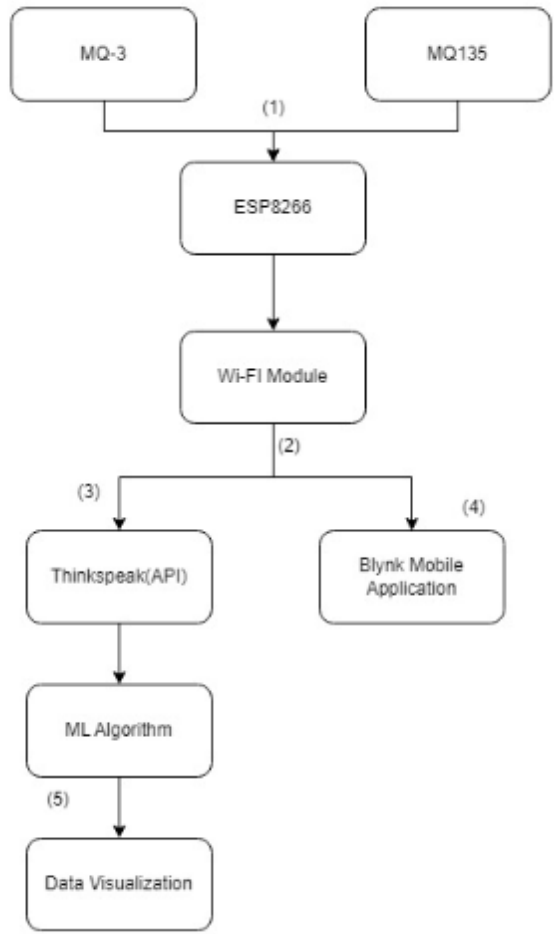

Figure 1: Flow diagram.

1) Input from MQ3, MQ135 is taken in the format of analog and the send to the ESP8266.

2) The Wi-Fi module conneccted to the nearby Wi-Fi sends the data to ThingSpeak and Blynk IoT platform.

3) The raw data from the sensors is sent to the ThingSpeak.

4) The calculated ppm values are sent to the Blynk IoT mobile application.

5) The data collected on ThingSpeak is exported as csv file and processed through the ML algorithm for

data visualization.

## CIRCUIT DESIGN

The sensors selected for the system were the MQ 135 gas sensor for volatile organic compounds (VOCs) and the MQ 3 gas sensor for alcohol. The sensors were calibrated by exposing them to known levels of pollutants and adjusting the readings to match the expected values. The hardware design also consisted of an ESP8266, Wi-Fi module, MQ 135, MQ 3 gas sensors, an Arduino microcontroller, and a power source.

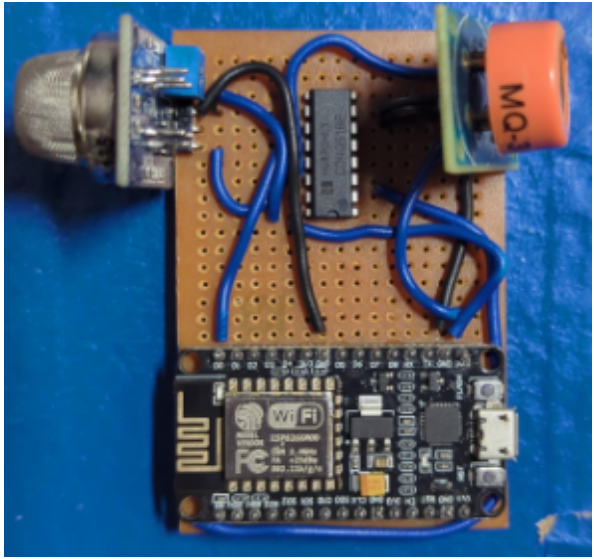

Figure 2: Circuit Design

The sensors were connected to the microcontroller, and the ESP8266 was used to establish a Wi-Fi connection for data transmission. The software development involved programming the microcontroller using Arduino IDE.

The code was designed to read data from the sensors, process it, and transmit it wirelessly to a cloud-based server. The system collects and stores data from the sensors at regular intervals. The collected data were analyzed using machine learning algorithms to predict future pollution levels.

## ALGORITHM

The below algorithm is followed to collect data from the sensors.

1. Define the Blynk credentials, WiFi credentials, and other variables required for the code.

2. Setup the serial communication and the Blynk connection using Blynk.begin().

3. Set up the timer to run a function to send data to ThingSpeak every second.

4. Connect to the WiFi network using WiFi.begin() and wait until the connection is established.

5. Define the changeMUX function and set the MUX_A pin as output.

6. In the loop, run the Blynk and timer functions, and read the sensor data from the analog pin A0.

7. Calculate the sensor value 1 (ppm (parts per million)) value for the sensor data using a formula.

8. Read the sensor data from A0 for a total of six times, and take the average of these readings to get the

sensor value 0.

9. Change the MUX_A pin to HIGH, and read the sensor data from A0 for another six times, and take the average of these readings to get the sensor value 1.

10. Connect to ThingSpeak using the WiFiClient object.

11. Build the request string with the ThingSpeak API key and field values (sensorValue0 and sensorValue1) and send the GET request using the HTTPClient object.

12. Delay for a second before running the loop again.

13. Define the function to be called by the timer to send data to ThinkSpeak.

14. Change the MUX_A pin to LOW and read the sensor data from A0.

15. Calculate the ppm value for the sensor data using a formula.

16. Change the MUX_A pin to HIGH and read the sensor data from A0 for a total of six times, and take the average of these readings to get the sensor value 2.

17. Write the sensor value 1 and sensor value 2 to virtual pins V 1 and V 2 respectively using Blynk.virtualWrite().

## RESULT AND DISCUSSION

The below figures represent the data interpretation in BLYNK and ThingSpeak which are connected with the help of Unique key.

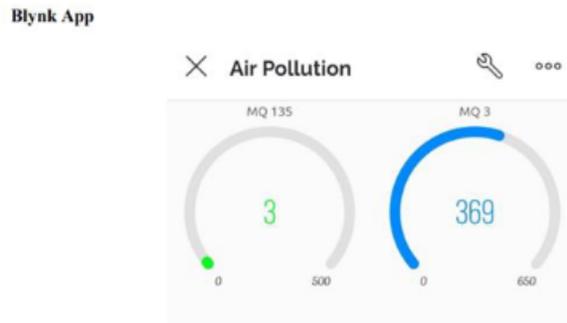

Figure 3: Data representation in Blynk Application

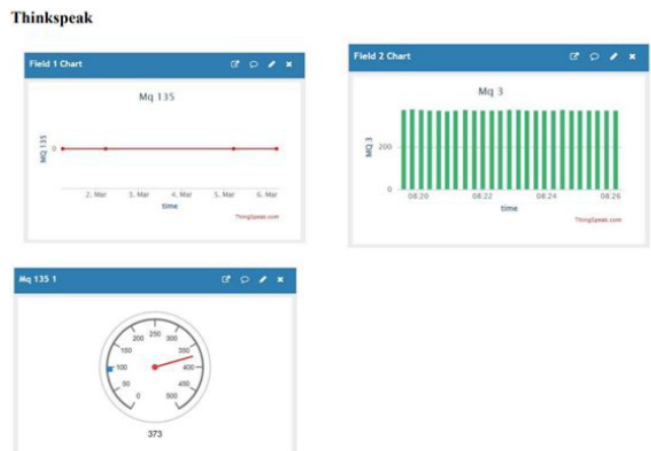

Figure 4: Data Stored in ThinkSpeak platform

In Figure 3, the green coloured meter represents the sensor readings of MQ135 sensor and the blue coloured meter represents the readings of MQ3 sensor.

The dataset used in this code is the"city_day.csv" dataset, which contains daily air quality data for multiple cities in India from 2015 to 2020. The dataset has been sourced from the Central Pollution Control Board (CPCB) of India.

The dataset has 15 columns, which include the following features: City, Date, PM2.5, PM10, NO, NO2, NOx, NH3, CO, SO2, O3, Benzene, Toluene, Xylene, and AQI.

The target variable in this code is AQI, which stands for Air Quality Index, and is a measure of the air quality based on the concentration of pollutants present in the air. The dataset contains daily air quality readings for different cities, and hence, has a large number of records.

The dataset also has missing values, which have been dropped in this code. Additionally, some of the features in the dataset show skewness, which can impact the performance of machine learning models. The code in this example uses three different regression models - Random Forest, Linear Regression, and Decision Tree - to predict AQI based on the concentration of pollutants.

The models have been evaluated using various metrics such as mean absolute error, root mean squared error, root mean squared logarithmic error, and R-squared score.

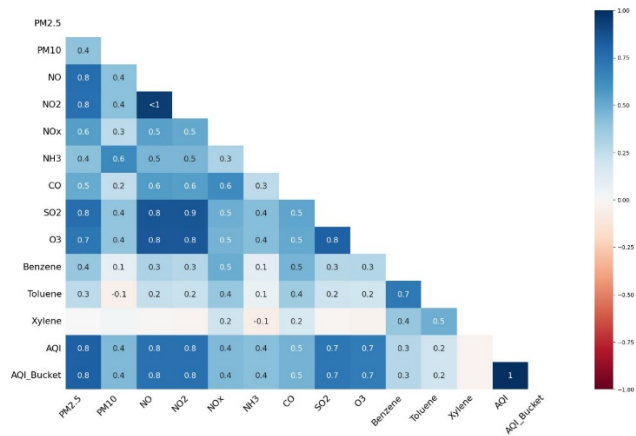

Figure 5: Correlation heat map of AQI with other pollutants.

In Figure 5 it can be seen that there are blanks in the correlation heat map data, which there are missing values for one or both of variables being compared. It could be said that there is so missing data of Xylene in the given data set. It can be seen that toluene and PM10, NH3 and xylene are inversely proportional to each other.

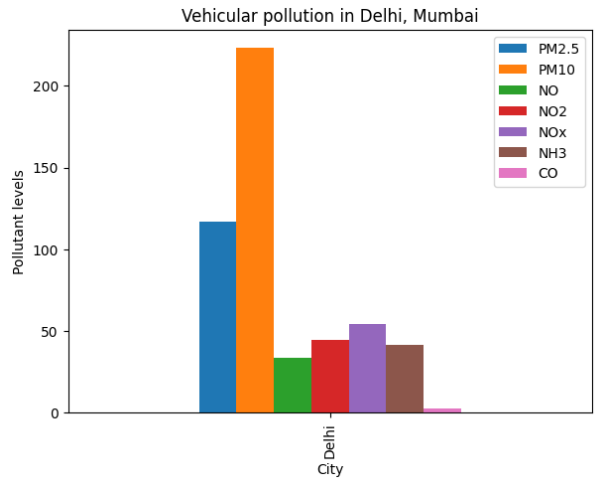

Figure 6: Vehicular pollution in Delhi, Mumbai

In Vehicular pollution the level of PM is high followed by PM2_5 from Figure 6

In industrial Pollutants the level of O3 is very high followed by toluene from Figure 7 in the cities of Mumbai and Delhi. It's important to note that ozone is a secondary pollutant, which means it is not emitted directly into the air, but rather forms because of chemical reactions between other pollutants (VOCS and Nitrogen oxides).

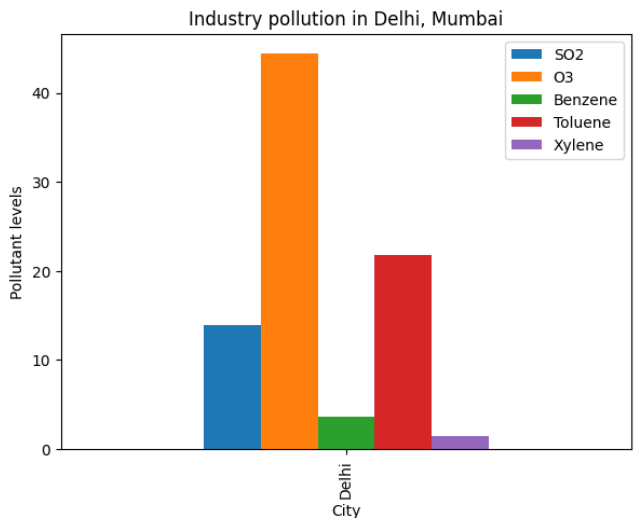

Figure 7: Industry Pollution in Delhi, Mumbai

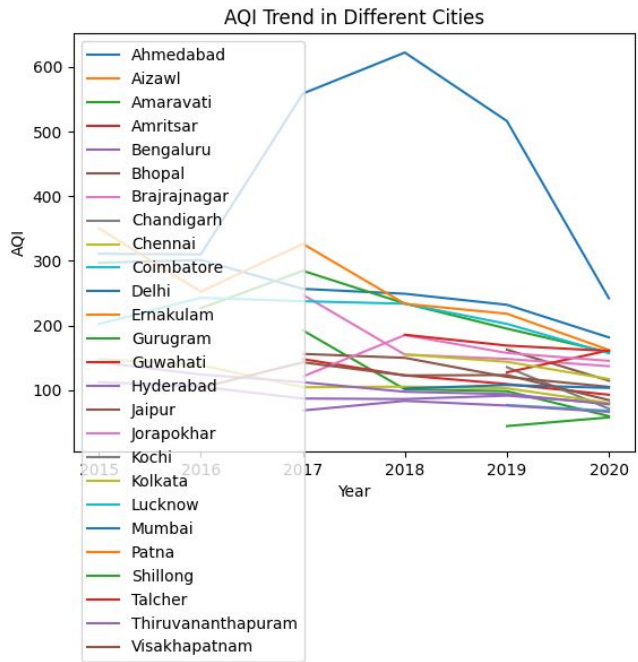

Figure 8: AQI trends in different Cities

As we see in Figure 8 there is no regular pattern of AQI index, there are so many external factors which affect the AQI of a particular city. It has been recorded that the highest Aqi was in Ahmedabad in the year 2018. The least AQI recorded was in Shillong.

From figure 8, The top 9 cities which have high industrial pollutants were noted in Patna, Delhi, Kolkata, Amritsar, Visakhapatnam, Amaravati, Hyderabad, Gurugram, Chandigarh.

The top 9 cities that have high vehicular pollutant was noted in Delhi, Patna, Amritsar, Visakhapatnam, Gurugram, Kolkata, Hyderabad, Chandigarh, Amaravati from Figure 9.

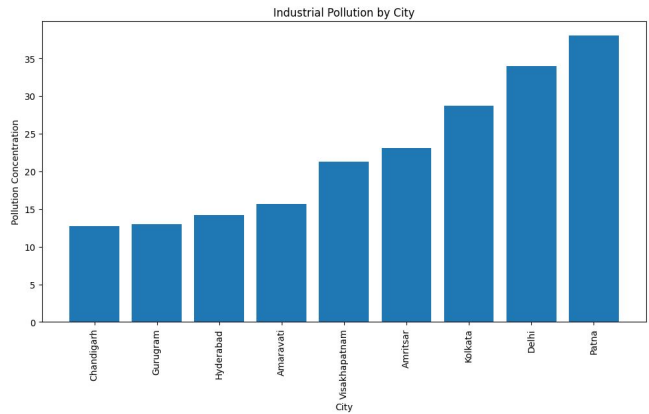

Figure 9: Industrial Pollution by city

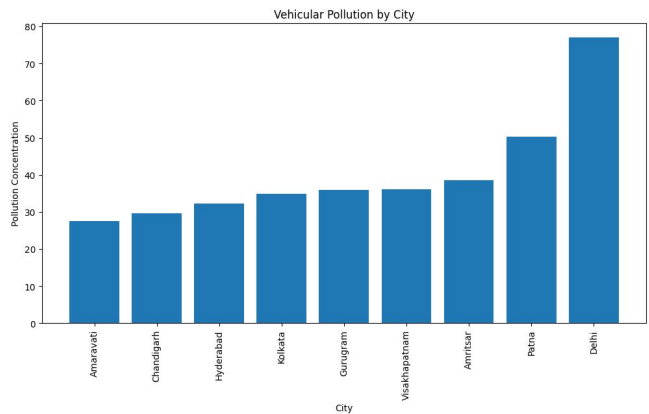

Figure 10: Vehicular Pollution by city

To address the issue of these high pollutant concentrations some solutions are discussed.

We used three regression models: Random Forest Regression, Linear Regression, and Decision Tree Regression. To evaluate the performance of these models we calculated Mean Absolute Error (MAE), Root Mean Squared Error (RMSE), Root Mean Squared Logarithmic Error (RMSLE), and R squared (R2).

MAE measures the average absolute difference between the predicted and actual values, whereas RMSE measures the square root of the average of the squared differences between the predicted and actual values.

RMSLE measures the root mean squared logarithmic error between the predicted and actual values, and it is used when the target variable has a large range of values. R-squared is a statistical measure that represents the proportion of variance in the dependent variable that is explained by the independent variables in the regression model.

Based on the results of these metrics we can compare the performance of the different models and choose the best one for our dataset. The model with the lowest MAE, RMSE, and RMSLE and the highest R-squared value is the best model for the dataset.

By adding SMOTE (Synthetic Minority Over-sampling Technique) to the regression models we can com-

pare the results, and we can conclude about using smote in Regression models.

| Model | MAE | RMSE | RMSLE | $R^2$ |
|---|---|---|---|---|
| Random Forest Regression | 14.78 | 21.59 | 0.16 | 0.94 |
| Linear Regression | 18.98 | 26.46 | 0.19 | 0.92 |
| Decision Tree Regression | 20.81 | 30.33 | 0.16 | 0.89 |

Table 1: For prediction of AQI: Without SMOTE.

| Model | MAE | RMSE | RMSLE | $R^2$ |
|---|---|---|---|---|
| Random Forest Regression | 15.02 | 22.00 | 0.17 | 0.94 |
| Linear Regression | 20.6 | 26.95 | 0.21 | 0.91 |
| Decision Tree Regression | 20.48 | 30.21 | 0.22 | 0.89 |

Table 2: For prediction of AQI: With SMOTE

From the above table 1 and 2 provides gives us the results of various metrics with three different regression models with smote and without smote, compared to other regression models we can say that Random Forest regression model given the best results without smote.

From these results we can say that using a smote didn't improve the results. Here smote failed because of the target, for predicting AQI smote should not be used.

| Model | Accuracy | F1 Score |
|---|---|---|
| Random Forest | 81 | 81 |
| Logistic Regression | 43 | 44 |
| Decision Tree | 70 | 70 |

Table 3: Prediction of AQI_Bucket : Without SMOTE

| Model | Accuracy | F1 Score |
|---|---|---|
| Random Forest | 80 | 80 |
| Logistic Regression | 40 | 41 |
| Decision Tree | 71 | 71 |

Table 4: Prediction of AQI_Bucket : With SMOTE.

By comparing Table 3 and 4 we can say that using smote given better results with Decision Tree classifier it is because AQI_Bucket has number of counts of moderate - 2521, satisfactory - 2079, Poor – 648, good - 454, very poor - 410, severe – 124, in this condition by using smote it improved the results. But overall Random Forest classifier gives the better results without smote.

| Model | Accuracy | F1 Score |
|---|---|---|
| SVM | 79 | 79 |
| KNN | 80 | 80 |
| Naïve Bayes | 72 | 73 |
| DNN | 79.2 | 78.3 |

Table 5: Prediction of AQI_Bucket

In Table 5 by comparing all the machine learning models like SVM (Support Vector Machines), KNN (K-Nearest Neighbors), Naïve Bayes, Random Forest and Deep Neural Network we can say that KNN performed better for this dataset.

## SOLUTIONS

To reduce the emissions of gases CO, NO2, O3, SO2, NH3, PM2_5, PM10, NH3 one can implement various scientific methodologies and can reduce them in many ways.

**a) Scientific Methodologies**

To lower the levels of the previously mentioned gases, a variety of scientific methods can be used.

- Catalytic converters: Used in cars to transform dangerous pollutants like CO, NO, andNO2 into less dangerous ones. To achieve this, exhaust gases are passed through a catalyst, which sets off a chemical reaction that transforms the harmful gases into less dangerous ones.

- Scrubbers: Before exhaust gases are discharged into the environment, pollutants are removed from them using scrubbers in factories and power plants. By doing this, it may be possible to lessen the emissions of hazardous gases like SO2, PM2.5, andPM10.

- Flue gas desulfurization: This method removes Sulphur dioxide from exhaust gases coming from power plants and other industrial operations. The SO2 is changed into a less dangerous compound that can be safely disposed of using a chemical procedure to accomplish this.

- Selective catalytic reduction: This procedure helps power plants and other industrial processes decrease their nitrogen oxide emissions. This is accomplished by injecting a reductant into the exhaust gases, typically ammonia or urea, which interacts with the NOx to transform it into harmless nitrogen and water vapor.

- Biofiltration: A natural method of purifying the air is known as biofiltration. By sending the air through a biofilter, which has microorganisms that convert the contaminants into harmless chemicals, this is accomplished.

- Carbon capture and storage: This procedure collects carbon dioxide from industrial processes and stores it in a secure area, like an underground storage facility. The amount of CO2 released into the atmosphere, which is a significant cause of climate change, can be decreased as a result. It's vital to remember that the most efficient method will depend on the precise source of the emissions and the surrounding environment. As a result, it's critical to do a thorough study of the problem and create a customized solution that considers all of the necessary variables.

**b) General Methods**

These gases are examples of air pollutants that can harm both the ecosystem and human health. The following are some ways to lower their levels.

- Cut back on pollution: Reducing the emissions of these gases is one of the best methods to lower.

  air pollution. Stricter emission regulations for factories and vehicles, encouragement of the use of renewable energy sources, and control of industrial processes that emit these gases can all help accomplish this.

- Encourage the use of public transportation: Since automobile emissions are a major source of air pollution, encouraging people to take the bus, walk, or bike instead of driving can help reduce those emissions.

- Enhance energy efficiency: Reducing emissions from power plants, another significant source of air pollution, can be accomplished by making structures and appliances more energy efficient.

- Plant trees: Trees are a useful instrument for decreasing air pollution because they absorb pollutants like carbon dioxide from the air. Urban tree planting can aid in lowering the airborne concentrations of these pollutants.

- Install air filters: By removing pollutants from the air and enhancing interior air quality, air filters can be installed in homes, workplaces, and public areas.

It's important to remember that these remedies need both individual and group efforts to be successful. Governments, corporations, and people must therefore cooperate to combat air pollution and safeguard both the environment and human health.

# CONCLUSION

An air quality monitoring system made of Arduino, a few sensors, and a multiplexer can be an effective and affordable solution for monitoring the air quality in a particular environment. The system can be easily customized and expanded with additional sensors depending on the specific needs of the application.

With the use of a multiplexer, multiple sensors can be connected to a single Arduino board, reducing the overall hardware cost and complexity. The collected data can be easily visualized and analyzed using software tools, allowing users to monitor and track air quality over time, detect potential issues and take necessary actions to improve the air quality.

Overall, this type of air quality monitoring system can be a useful tool for environmental monitoring, health management, and pollution control. By adding SMOTE (Synthetic Minority Over-sampling Technique) to the regression models we compared the results, and we came to a conclusion about using smote in Regression models.

One area of improvement could be in the calibration of the sensors, which could be optimized for specific environments or applications to improve accuracy. Additionally, machine learning algorithms could be used to identify patterns in the data and predict changes in air quality, allowing users to take preventive actions before air quality deteriorates. Another area of future work could be to improve the user interface and data visualization of the system, making it more accessible and easier to use for non-experts.

Overall, continued development and optimization of the system can help to increase its effectiveness and applicability in a variety of settings, from indoor air quality monitoring in homes and offices to outdoor pollution monitoring in urban environments.

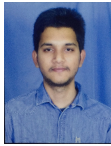

# AUTHORS PROFILE

**Hemanth Karnati**, He is a dedicated and enthusiastic 3rd-year student pursuing a Bachelor of Technology (B.Tech) degree in Computer Science and Engineering (CSE) at VIT Vellore. His passion for technology and innovation has driven him to explore various domains within the field. Hemanth's research work primarily revolves around the fields of IoT and ML. He has conducted extensive studies on leveraging IoT devices and networks to create intelligent systems that enhance efficiency, automation, and decision-making processes. His research work showcases his ability to apply ML algorithms and techniques to IoT data, enabling predictive analysis, anomaly detection, and optimization in diverse applications.